%% file: main.tex
\documentclass[a4paper,twoside]{article}

\usepackage{epsfig}
\usepackage{subcaption}
\usepackage{calc}
\usepackage{amssymb}
\usepackage{amstext}
\usepackage{amsmath}
\usepackage{amsthm}
\usepackage{multicol}
\usepackage{pslatex}
\usepackage{apalike}
\usepackage{algorithm2e}
\usepackage[bottom]{footmisc}
\usepackage{SCITEPRESS}

\usepackage{glossaries}
\glsdisablehyper
\usepackage{textcomp}
\usepackage[hidelinks]{hyperref}
\usepackage[none]{hyphenat}
\usepackage[capitalise]{cleveref}
\creflabelformat{equation}{#2#1#3}
\usepackage{tabularx}
\usepackage{booktabs}
\usepackage{gensymb}
\usepackage{adjustbox}

\newacronym{RL}{RL}{Reinforcement Learning}
\newacronym{MARL}{MARL}{Multi-Agent Reinforcement Learning}
\newacronym{FOV}{FOV}{Field of View}
\newacronym{PPO}{PPO}{Proximal Policy Optimization}
\newacronym{GAE}{GAE}{Generalized Advantage Estimation}

\begin{document}

\title{Aquarium: A Comprehensive Framework for Exploring Predator-Prey Dynamics through Multi-Agent Reinforcement Learning Algorithms}

\author{\authorname{Michael Kölle\sup{1}, Yannick Erpelding\sup{1}, Fabian Ritz\sup{1}, Thomy Phan\sup{2}, Steffen Illium\sup{1} and Claudia Linnhoff-Popien\sup{1}}
\affiliation{\sup{1}Institute of Informatics, LMU Munich, Munich, Germany}
\affiliation{\sup{2}Thomas Lord Department of Computer Science, University of Southern California, Los Angeles, USA}
\email{michael.koelle@ifi.lmu.de}
}


\input{content/0_abstract}

\onecolumn \maketitle \normalsize \setcounter{footnote}{0} \vfill

\input{content/1_introduction}
\input{content/2_preliminaries}
\input{content/3_related_work}

\input{content/4_approach}
\input{content/5_experimental_setup}
\input{content/6_results}
\input{content/7_conclusion}
\input{content/8_acknowledgements}
\bibliographystyle{apalike}
{\small
\bibliography{main}}

\end{document}

%% file: content/0_abstract.tex
\abstract{
Recent advances in Multi-Agent Reinforcement Learning have prompted the modeling of intricate interactions between agents in simulated environments. In particular, the predator-prey dynamics have captured substantial interest and various simulations been tailored to unique requirements. To prevent further time-intensive developments, we introduce Aquarium, a comprehensive Multi-Agent Reinforcement Learning environment for predator-prey interaction, enabling the study of emergent behavior. Aquarium is open source and offers a seamless integration of the PettingZoo framework, allowing a quick start with proven algorithm implementations. It features physics-based agent movement on a two-dimensional, edge-wrapping plane. The agent-environment interaction (observations, actions, rewards) and the environment settings (agent speed, prey reproduction, predator starvation, and others) are fully customizable. Besides a resource-efficient visualization, Aquarium supports to record video files, providing a visual comprehension of agent behavior. To demonstrate the environment's capabilities, we conduct preliminary studies which use PPO to train multiple prey agents to evade a predator. In accordance to the literature, we find Individual Learning to result in worse performance than Parameter Sharing, which significantly improves coordination and sample-efficiency.
}

%% file: content/1_introduction.tex
\section{INTRODUCTION}
\label{sec:introduction}
%Importance of RL
\gls*{RL} has emerged as a pivotal paradigm to train intelligent agents for sequential decision-making tasks in many domains, ranging from games, robotics to finance and healthcare~\cite{mnih_2015_atari,silver_2018_alphazero,haarnoja_2019_locomotion,vinyals_2019_alphastar}.
%. With its foundation rooted in the principles of reward maximization, \gls*{RL} has garnered substantial attention due to its applicability 

The ability of \gls*{RL} agents to learn from interacting with a problem and adapt their behavior to optimize for long-term benefit positioned \gls*{RL} as a promising approach to realize decision-making systems~\cite{RL_sutton_barto}.
%Importance of MARL
%Recently, the efficacy of \gls*{RL} techniques has triggered extensive research in understanding and modeling intricate interactions between multiple agents within an environment~\cite{wang2022modelbased}. This subdomain, known as 
\gls*{MARL} revolves around the dynamics of learning in the context of other agents and addresses scenarios like collaborative robotics, social dilemmas, and strategic games, where agent interactions are crucial for success~\cite{marl-book}.
%the study of how intelligent agents learn and adapt their strategies in response to the actions of other agents. \gls*{MARL}

%Importance of Pred-Prey
%Among the multiple applications of \gls*{MARL}, the investigation of predator-prey dynamics has garnered considerable attention.
Predator-prey domains are widely used to analyze aspects of agent cooperation, competition, adaptation, and learning~\cite{diz-pita_predatorprey_2021,li_predator-prey_2023}.
Rooted in ecological studies, this scenario models the interaction between pursuing predators and evasive prey, allowing to study a wide range of technical, societal and ecological aspects of multi agent systems.
%Past Studies
%Many studies have been carried out to systematically assess and analyze the behavior of predator and prey agents in such settings by investigating the interaction dynamics.
%These studies investigated a range of intriguing phenomena within such environments, shedding light on the dynamics that emerge from interactions between predator and prey agents. 
%The investigated phenomena in the predator-prey setting include 
For example, one line of work first found emergent swarming and foraging behavior among learning prey agents~\cite{hahn_emergent_2019,hahn_foraging_2020} and later showed that swarming is a nash equilibrium under certain conditions~\cite{hahn_nash_2020}.
Subsequently, sustainable behaviour of single learning predator agents~\cite{ritz_towards_2020} as well as herding and group hunting of multiple learning predator agents was achieved~\cite{ritz_sustainable_2021}.
Swarming and group hunting have prevailed at various places in nature, thus it is highly interesting to study the emergence of such phenomena among artificial agents.

While predator-prey scenarios are common in the RL community, studies have mostly been conducted in different environment implementations. 
Yet, developing RL environments is time-intensive and error-prone: abstraction and granularity have to be balanced, a trade off between simulation precision and computational speed has to be found, the choice of algorithm should not be restricted and reproducibility has to be ensured.
%However, designing environments that accurately capture emergent phenomena while remaining adaptable and  is essential for advancing our understanding of complex multi-agent systems.

\begin{figure}[!ht]
    \centering
    \includegraphics[width=\columnwidth]{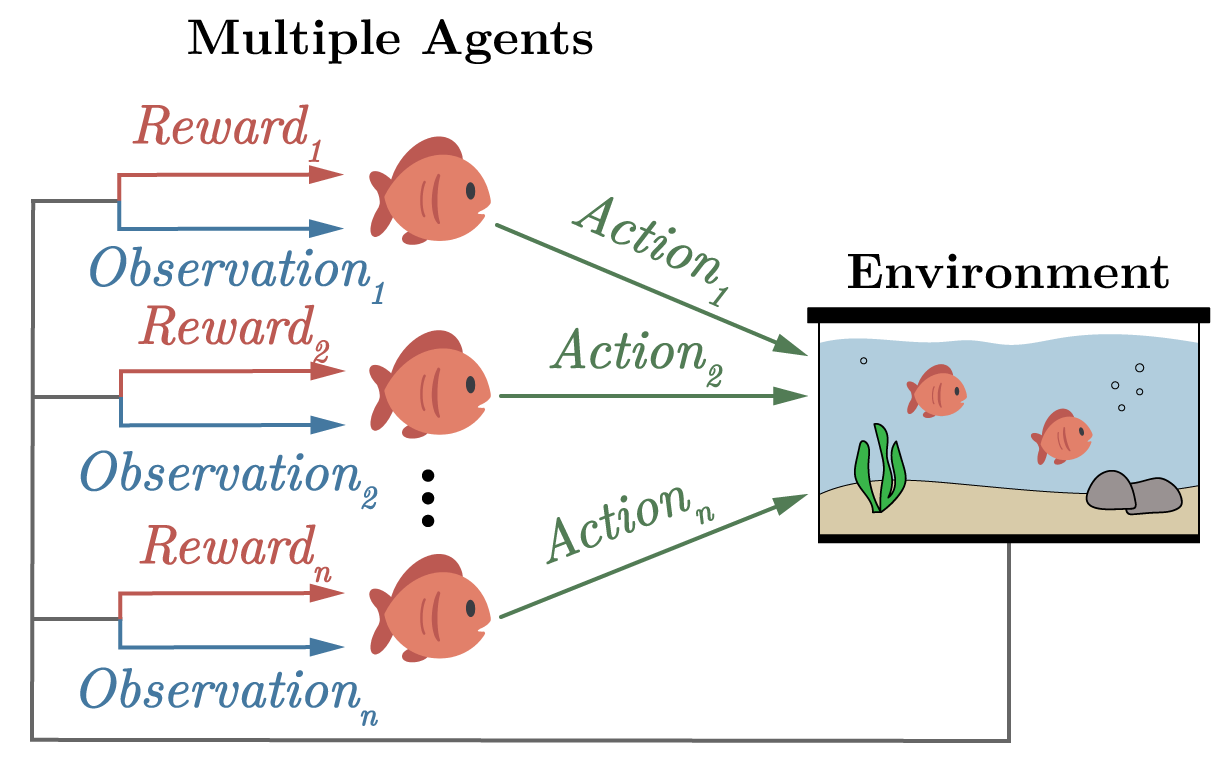}
    \caption{The Multi-Agent Reinforcement Learning Cycle (c.f.~\cite{terry_multi-agent_2023}). Within the aquarium environment of \(n\) agents, each optimizes its policy to maximize individual rewards, while concurrently influencing the observations and rewards of others.}
    \label{fig:MARL}
\end{figure}

Thus, we present a standardized environment for future research on predator-prey scenarios (see \cref{fig:MARL}) that builds upon the environments used in~\cite{hahn_emergent_2019,hahn_foraging_2020,hahn_nash_2020,ritz_towards_2020,ritz_sustainable_2021}.
Our implementation offers a seamless integration of the \textit{PettingZoo} framework~\cite{terry_pettingzoo_2021}, allowing a quick start with proven \gls*{MARL} algorithm implementations.
As our implementation is fully customizable, it offers a flexible and re-usable base to efficiently explore a variety of scenarios, e.g. regarding population dynamics, social dilemmas, sustainability, and (self-) organisation. Our contributions are:
\begin{enumerate}
    \item An overview of predator-prey environments used in the (MA)RL community.
    \item A unified yet customizable environment that covers all identified aspects and is compatible to the proven MARL algorithm implementations of the PettingZoo framework.
    \item Preliminary experiments reproducing emergent behaviour of learning agents and demonstrating the scalability of modern MARL paradigms in our environment.
\end{enumerate}

% \begin{figure}[htb]
%     \centering
%     \includegraphics[width=\columnwidth]{images/RL_big_font.png}
%     \caption{The Reinforcement Learning Cycle c.f.~\cite{terry_multi-agent_2023}. The agent selects an action, altering the environment's state, and receives immediate feedback: a reward and a new observation. These inform the RL algorithm to update the policy, determining the subsequent action.
% }
%     \label{fig:RL}
% \end{figure}

%Thus, novel research investigations can be sped up due to the presence of a standardized environmental framework. 
%Furthermore, the exploration of \gls*{MARL} predator-prey dynamics involving independent learning is examined as an illustrative instance of how this environment can facilitate novel research investigations.

Our work is structured as follows. In the following \cref{sec:pred_prey_bg}, we describe the predator-prey scenario. In \cref{sec:related-work}, we review existing applications of the predator-prey scenario to identify all key components and concepts that are required for thorough studies. In the subsequent \cref{sec:approach}, we explain how we incorporated these components and concepts in our \textit{Aquarium} environment, which forms the basis of our experimental framework. In \cref{sec:experimental-setup}, we describe our experimental setup, and then provide and discuss the according results in \cref{sec:results}. We conclude in \cref{sec:conclusion} with a summary of key insights, current limitations and suggestions for future research.

%% file: content/2_preliminaries.tex
\section{PREDATOR-PREY SCENARIO} 
\label{sec:pred_prey_bg}

Population dynamics (biomathematics) study how single and multiple species coexist and interact in the same habitat~\cite{diz-pita_predatorprey_2021}. 
Here, the predator-prey scenario models the dynamics between two types of organisms: predators and prey. 
Predators are organisms that hunt and consume other organisms.
Prey are organisms that try to evade and survive.
In nature, some examples are lion and zebra, fox and rabbit, or shark and fish.

Independent of each other, \textit{Volterra} and \textit{Lotka} initiated investigations on the ecological predator-prey scenario in the 1920s~\cite{lotka_1925,volterra_1926}.
Both formulated a pair of first-order nonlinear differential equations that describe the dynamics of predator-prey interactions within an ecological system.
These involve two primary variables and capture the instantaneous growth rates of both populations (prey $x$, predator $y$):
\begin{equation}\label{eqn:pred_prey}
\frac{dx}{dt} = \alpha \cdot x - \beta \cdot x \cdot y 
\quad\text{and}\quad 
\frac{dy}{dt} = \delta \cdot x \cdot y - \gamma \cdot y
\end{equation}
where $t$ is the time, $\alpha$ is the prey reproduction rate, $\beta$ is the predation rate, $\delta$ is the predator reproduction rate and $\gamma$ is the predator mortality rate.

The system dynamics are characterized as follows: 
when the predator population increases, this exerts pressure on the prey population, leading to a decline of available prey.
As a result, the predator population may then also decline due to reducing food supply. 
With fewer predators, the prey population can recover, initiating a new cycle of population fluctuations, which leads to a balance in the ecosystem.

Furthermore, the predator-prey relationship can lead to co-evolution between the two groups.
Through continuous interactions and adaptations, both populations may undergo reciprocal evolutionary changes.
The prey may develop defensive mechanisms, such as camouflage, warning signals, or toxins, to deter predation.
In response, the predators may evolve better hunting tactics, specialized adaptations, or improved sensory capabilities to overcome these defenses. 
%This coevolutionary arms race can give rise to fascinating adaptations in both predator and prey species.

%Moreover, the predator-prey relationship holds significant importance in maintaining biodiversity and promoting the well-being of ecosystems. The predators play a crucial role in regulating the population levels of certain prey species, thereby preventing any species from becoming too dominant and allowing other species to thrive in the ecosystem. This diversity supports greater stability and resilience in the face of environmental changes, making the ecosystem more robust.

Understanding the predator-prey relationship is essential for conservation efforts and sustainable management of ecosystems.
By studying and conserving these interactions, scientists, and ecologists can better understand the elaborate system of intertwined life dynamics and the delicate balance that sustains natural environments.
With increasing computational resources, predator-prey simulations have become a common environment for Single- and Multi-Agent \gls*{RL} algorithms, which has in fact been advocated 25 years ago~\cite{grafton_how_1997}.

{}

%% file: content/3_related_work.tex
\section{RELATED WORK} \label{sec:related-work}
We aim to provide a versatile \gls*{RL} environment with all key features of previous research.
To identify these, we review applications of (MA-) RL to predator-prey scenarios in the following.

The first aspect is the spatial representation.
Discrete, two-dimensional predator-prey simulations for multi-agent systems have been around for more than 20 years~\cite{stone_2000_mas}, sometimes also referred to as \textit{pursuit-evasion}.
In discrete simulations, agent positions and movement are restricted to grid cells.
Due to their good scalability, they are often used as a preliminary test bed for cooperation~\cite{gupta_2017_marl} and resilience~\cite{phan_2021_resilience} in \gls*{MARL}, as well as to study the dynamics of large populations with up to a million agents~\cite{yang_2018_million}.
The latter also observed the emergence of Lotka-Volterra cylces.
However, we refrain from modeling discrete positions as such a coarse spatial representation prevents meaningful studies on swarming behavior: this requires precise steering to adjust orientation, distance and alignment~\cite{reynolds1999steering}.
To enable agents to move accordingly, a continuous spatial model is required.

The next key aspect is the number of dimensions.
Modeling three-dimensions~\cite{berlinger_2021_3D} comes at the cost of increased mathematical and conceptual complexity.
In fact, a continuous, two-dimensional, edge-wrapping plane has been shown shown to be sufficient to study complex agent interactions:
By training prey with the RL algorithms DQN and DDPG to survive as long as possible,~\cite{hahn_emergent_2019} observed the emergence of flocking behavior in the presence of predators as described by ~\cite{reynolds_flocks_1987}.
Follow-up research by~\cite{hahn_nash_2020} showed that swarming can be a sub-optimal Nash equilibrium in predator-prey scenarios and illustrated how (not) forming a swarm puts the prey into a social dilemma, demonstrating that two dimensions are sufficient to explore the complexity of the interplay between individual and collective behaviors in swarming dynamics.
In a simulation with similar characteristics,~\cite{huettenrauch_2019_swarms} propose a MARL variant of the RL algorithm DDPG.
Their agents communicate in local neighborhoods, e.g. to exchange information about targets to be localized, and use an efficient representation of local observations to improve scaling.

Despite a suitable spatial representation, the aforementioned environments do not consider collisions and allow agents to overlap, which is the third key aspect.
Missing collisions make it unnecessary for prey agents to keep distance to each other and greatly simplifies to escape from predators.
The work of~\cite{ritz_towards_2020} added elastic collisions and a moment of inertia to the environment of~\cite{hahn_emergent_2019} and trained an RL predator to capture prey.
In a two-phase training approach, the predator adapted towards a balanced strategy to preserve the prey population.
Subsequently,~\cite{ritz_sustainable_2021} found multiple learning predators to learn sustainable and cooperative behavior amid challenges such as starvation pressure and a tragedy of the commons.
Environment parameters such as edge-wrapping, agent speed, or the observation radius to significantly impacted their results.

Another line of research uses a two-dimensional plane with collision physics and additional obstacles:
\cite{mordatch_2018_language} study the emergence of communication amongst learning agents and~\cite{lowe_2017_maddpg} propose the MARL algorithm MADDPG, which achieves significantly better results than purely decentralized actor-critic algorithms.
This suggests that communication and organization are easier to learn centrally when scenarios require coordination. 
However, their environment lacks edge-wrapping.
This allows agents to (theoretically) move to infinity and for practical reasons, agents need to be stopped from doing so.
Thus, the concept of landmarks to which agents shall (always) navigate to is used, which limits the applicability to study swarm behavior.

However, all aforementioned environments lack one key feature that has received little attention in simulations so far but could have significant impact on the preys' formation and the predators' hunting: agent \gls{FOV} (see \cref{sec:Vision}). This would allow to study the \textit{Many Eyes Hypothesis}~\cite{olson_2015_vigilance} with RL.

In summary, a unified environment should
\begin{enumerate}
    \item enable continuous agent navigation in a unbounded 2D plane through edge-wrapping similar to~\cite{hahn_emergent_2019},
    \item use a thorough physics model, e.g. with a moment of inertia and collisions between agents and obstacles, similar to~\cite{lowe_2017_maddpg,ritz_towards_2020},
    \item allow a flexible environment parameterization similar to~\cite{ritz_sustainable_2021} and
    \item model agent FOV.
\end{enumerate}
We realize all these key aspects in our \textit{Aquarium} environment, which we describe in the following.

%% file: content/4_approach.tex
\section{AQUARIUM} \label{sec:approach}
We introduce \textit{Aquarium}, a comprehensive and flexible environment for MARL research in predator-prey scenarios.
By making key parameters accessible, we support for the examination of a wide array of research questions. 
Our goal is to eliminate the necessity for researchers to rebuild basic dynamics, which is often a significantly time-consuming task, thereby affording more opportunities for the methodical analysis of relevant factors.
Moreover, employing a unified simulation platform eases the reproduction of previous results and strengthens the reliability of performance comparisons between different methodologies via a stable foundation.
To include the collective effort put into existing RL libraries, we implement the \textit{PettingZoo} interface~\cite{terry_pettingzoo_2021}, which enables out-of-the-box support for CleanRL, Tianshou, Ray RLlib, LangChain, Stable-Baselines3, and others.
The project is open-source, distributed under the MIT license, and available as a package on PyPI \footnote{\href{https://github.com/michaelkoelle/marl-aquarium}{https://github.com/michaelkoelle/marl-aquarium}}. %The nomenclature of our environment draws from its inherent configuration, where sharks are identified as predators and fish as prey, symbolizing the embedded predatory-prey interactions within the simulated scenarios.

%\subsection{Prerequisites}\label{sec:ppa_motivation}
% To undertake a research study within the domain of MARL, it is imperative to establish an environment, as necessitated for all the studies introduced in Section~\cref{sec:related-work}. Therefore, the aim of this study was to develop a predator-prey environment characterized by flexibility and re-usability in order to facilitate future MARL research on predator-prey problems. The research of multiple common predator-prey phenomena, such as chasing, evading, and capturing would be facilitated while exposing key parameters to enable studying a wide range of research questions. Consequently, researchers are alleviated from the requirement of reconstructing fundamental dynamics, which can be time-intensive. This, in turn, grants them more opportunity to systematically analyze factors of interest. Moreover, employing a shared simulation platform facilitates the replication of previous outcomes and fosters more substantial performance comparisons between approaches. 

\subsection{Dynamics and Perception}\label{sec::ppa_environment}

\begin{figure}[htbp]
    \centering
     \subfloat[Simple Visualization.]{%
        \includegraphics[width=0.47\columnwidth]{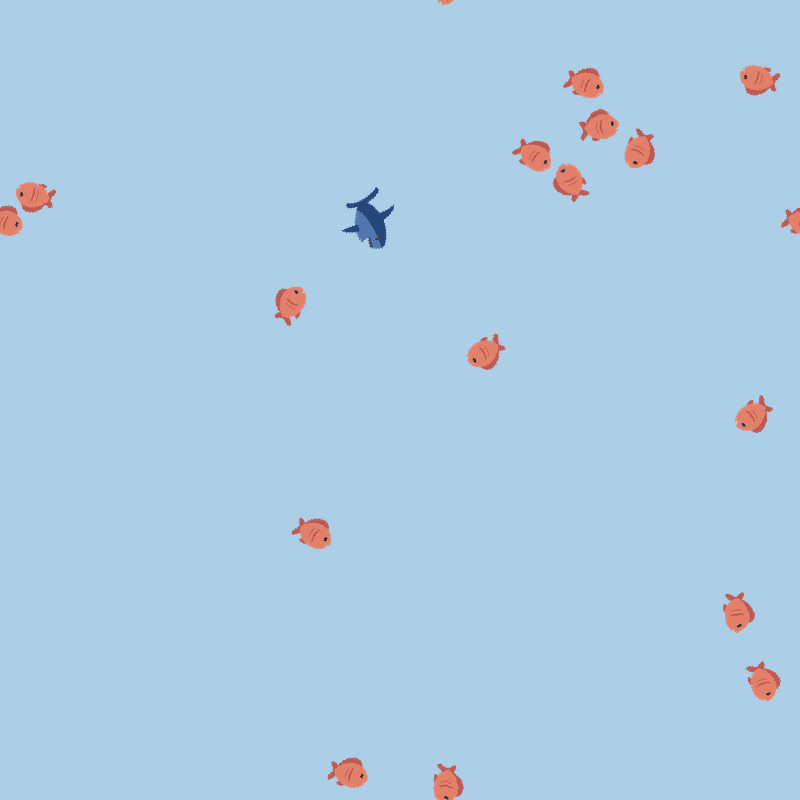}
        \label{fig:env_example_simple}
    }
    \hfill
    \subfloat[Visualization Including Force Vectors and Cones.]{%
        \includegraphics[width=0.47\columnwidth]{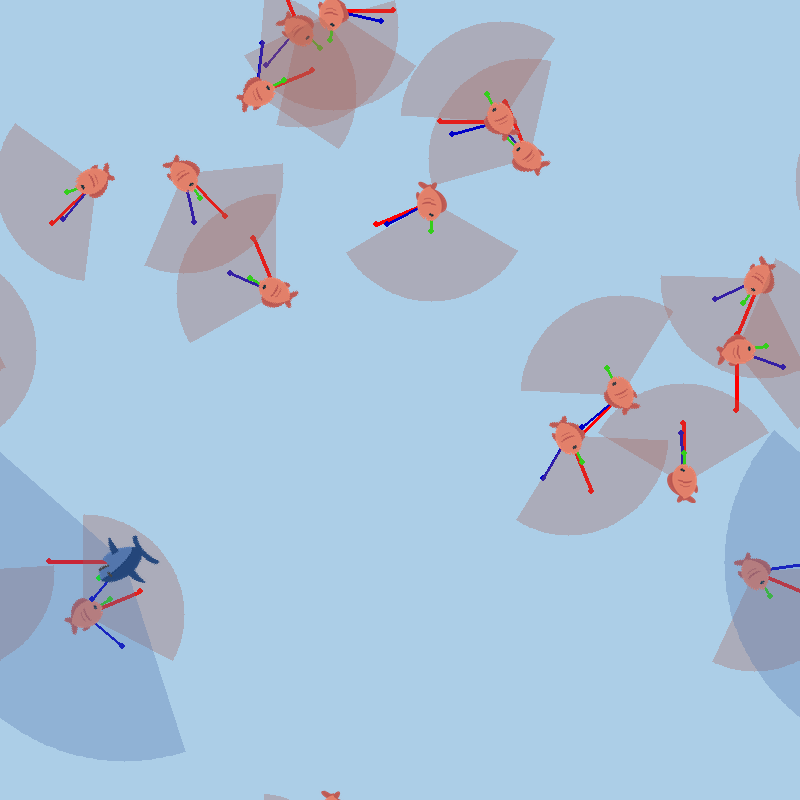}
        \label{fig:env_example_cone}
    }
    \hfill
    \vspace{5pt}
    \caption{Examples of the Aquarium With 16 Prey (Fishes) and One Predator (Shark) Agent.}
    \label{fig:env_example}
\end{figure}

The agents in our environment traverse a continuous two-dimensional toroidal space, enabling seamless movement across its boundaries (\cref{fig:env_example_simple}). Detailed computations for this toroidal configuration are explored in the following sections.

The environment supports multiple instances of two agent types: prey, visually represented as fish, and predators, visually represented as sharks. The actual position of agents is marked by central point within these representations. Prey agents can replicate after a specified survival duration, with the process capped upon reaching a predefined maximum prey count. An episode terminates either after a predetermined number of time steps or when all agents of one type are eliminated, with predators also susceptible to starvation after a set number of unsuccessful hunting steps. Agents are characterized by attributes like mass, position, velocity, and acceleration, and utilize essential vectors, namely \textit{position}, \textit{velocity}, and \textit{acceleration} vectors, for navigation.

\subsubsection{Movement}\label{sec:Agent_Movement}

\begin{figure}[htbp]
    \centering  \includegraphics[width=0.48\columnwidth]{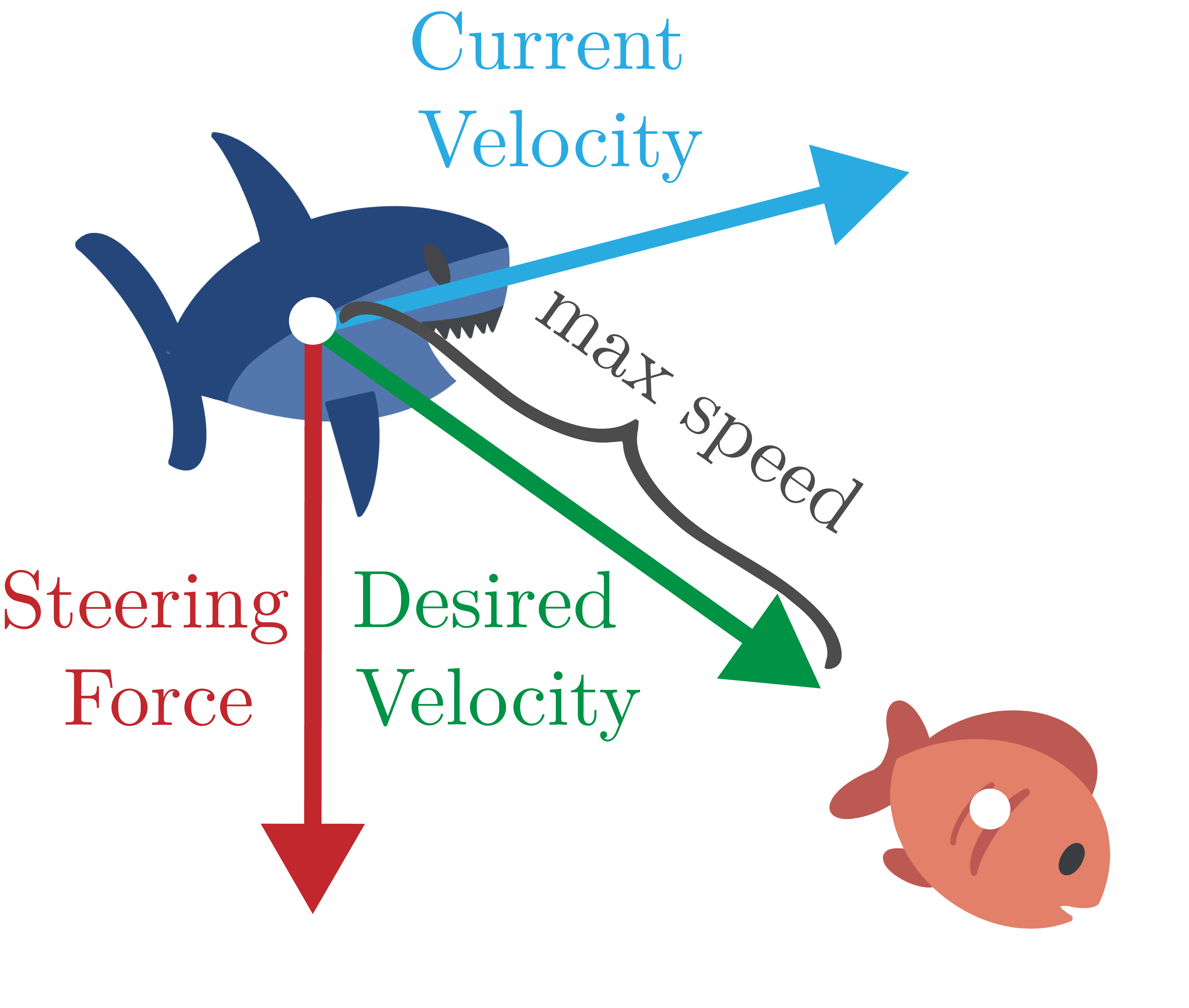}
    \caption{Calculation of the Steering Force. A limited fraction of the steering force is applied to smoothly transition from the current to the desired velocity, which is capped by the agent's maximum speed.
    }
    \label{fig:steering}
\end{figure}

In the context of a predator-prey scenario, an agent's primary objective involves initiating a directional movement, which necessitates the manipulation of its velocity through the application of forces. At each time step, a steering mechanism, derived from the algorithm of~\cite{reynolds1999steering}, is employed to guide the agent towards a designated position within the environment (\cref{fig:steering}). This adaptation aligns the agent, for instance, a predator, and its velocity is radially aligned towards its designated target, a prey. 

The desired velocity $D$, a vector pointing from the agent to the target, is constrained in magnitude by the agent's maximum attainable speed $s \in \mathbb{R}$. The steering vector $F$ is the difference between the desired velocity $D$ and the agent's current velocity $V$ and is limited by the agent's maximum steering force $f \in \mathbb{R}$ (\cref{eq:steering_force}).

\begin{equation}\label{eq:steering_force}
F = \begin{cases} D-V & \text{if } \|D - V \| \leq f \\ \frac{D -V}{\|D -V\|} \cdot f & \text{otherwise} \end{cases}
\end{equation}

After the calculation of the steering force, the agent's new acceleration vector $A'$ is calculated using the agent's current acceleration vector $A$, the steering force $F$, the agent's mass $m \in N$, and the maximum magnitude $k \in \mathbb{R}$ of the final acceleration vector (\cref{eq:acceleration}). 

\begin{equation}\label{eq:acceleration}
A' = \frac{A + \frac{F}{m}}{\|A + \frac{F}{m}\|} \cdot k,
\end{equation}

The new acceleration vector $A'$ is then added to the agent's current velocity vector $V$ to compute the new velocity vector $V'$, which is then limited by the agent's maximum speed $s$.

\begin{equation}\label{eq:new_velocity}
V' = \begin{cases} V + A' & \text{if } \|V + A'\| \leq s \\ \frac{V + A'}{\|V + A'\|} \cdot s & \text{otherwise} \end{cases}
\end{equation}

To ultimately change the agent's position, the new velocity vector $V'$ is added to the agent's position vector $P$ (\cref{eq:agent_position}).

\begin{equation}\label{eq:agent_position}
P' = P + V'
\end{equation}

To increase the flexibility of the environment, the maximum magnitudes of the desired velocity $s$, the steering force $f$, and the acceleration vector $k$ can be manipulated, independently of the agent type. 

\subsubsection{Collisions}\label{sec:Agent_Collisions}

As a predator-prey setting consists of predators trying to capture their prey, interactions between agents are enabled in this environment. Every agent is equipped with a circular hitbox of a predetermined radius $r \in \mathbb{R}$ around the agent's center. Detecting collisions between two agents $A$ and $B$ involves calculating the distance between them using their positional vectors $P_A$ and $P_B$. If the computed distance $d$ is smaller than the the sum of the radii of both agents, i.e. $d < r_A + r_B$, agents $A$ and $B$ have collided. 

When a predator collides with a prey, the prey is considered captured and can be immediately respawned at a random position within the environment if configured. This ensures a constant number of prey throughout the entire episode. On the other hand, if two agents of the same type collide with each other, they bounce back in the opposite directions of the other agent.  Following a collision, the new velocity $V_{A}'$ of agent $A$ is calculated using its current velocity $V_A$ and the positional vectors of both agents $A$ and $B$ (\cref{eq:new_velocity_collision}). The new velocity $V_{B}'$ of agent $B$ is determined analogously.

\begin{equation}\label{eq:new_velocity_collision}
V_A' = V_A + \frac{P_B - P_A}{\|P_B - P_A\|}
\end{equation}

\subsubsection{Vision}\label{sec:Vision}

In an agent-based system, the actions of an agent are influenced by the information it perceives from its environment. Therefore, it is crucial to precisely define what the agent can see. In the default setup of this environment, each agent has complete visibility of all other agents, implying that it knows the exact positions of every agent. However, for a more realistic simulation, we can impose restrictions on the agent's view, rendering the environment partially observed. A widely-used approach involves establishing a limited viewing distance, and thus creating a circular vision field around the agent. Consequently, the agent can only perceive other agents situated within this predetermined viewing distance.

\begin{figure}[htbp]
    \centering
    \includegraphics[width=\columnwidth]{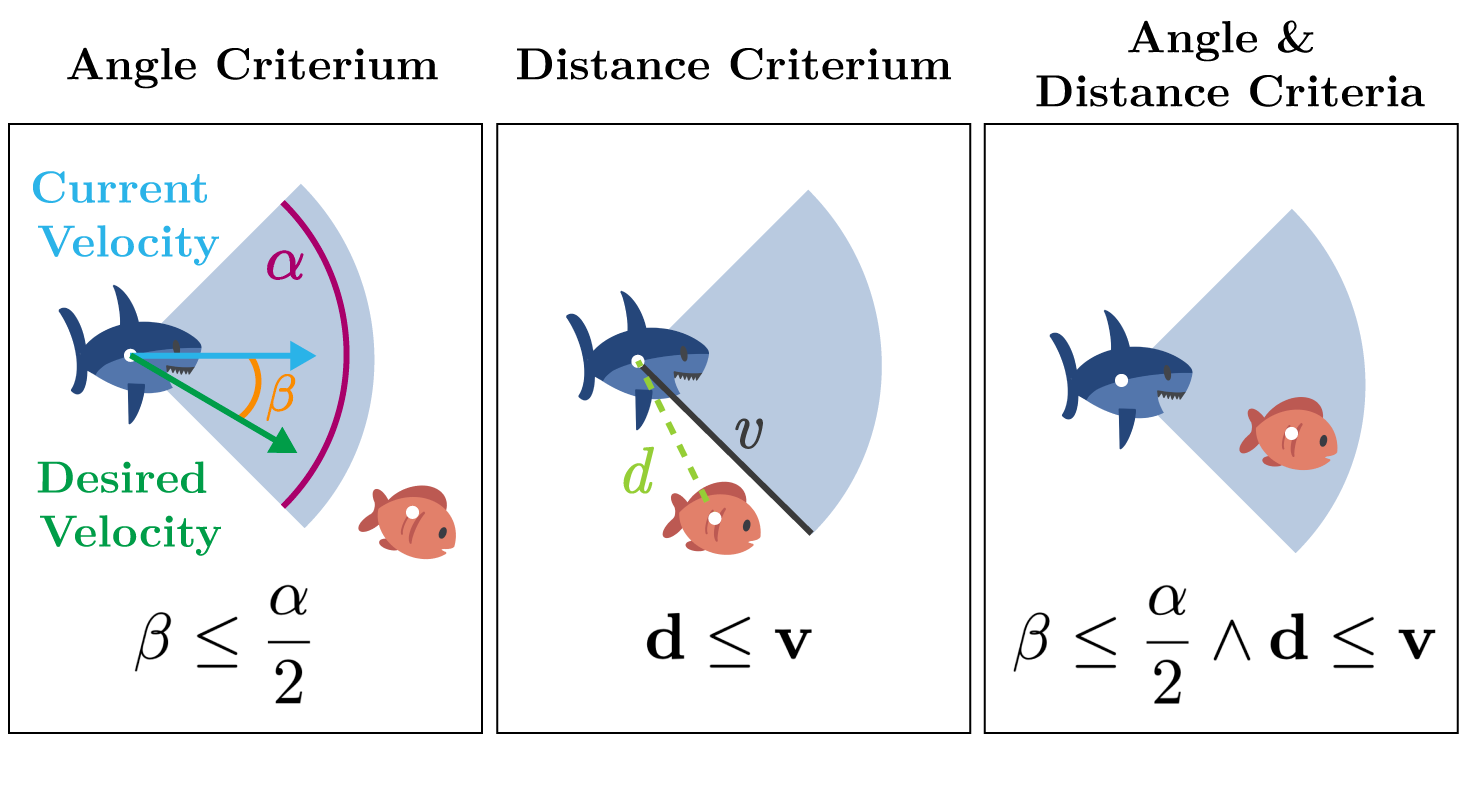}
    \caption{Angle and Distance Criteria. An agent is considered within the field of view only if both conditions are met: staying within a defined angular range, $\alpha$ (left), and not exceeding a specified distance, $v$ (middle). Both conditions are fulfilled (right).}
    \label{fig:cone_detection}
\end{figure}

\begin{figure}[htbp]
    \centering
    \subfloat[Field of View.]{%
        \includegraphics[width=0.47\columnwidth]{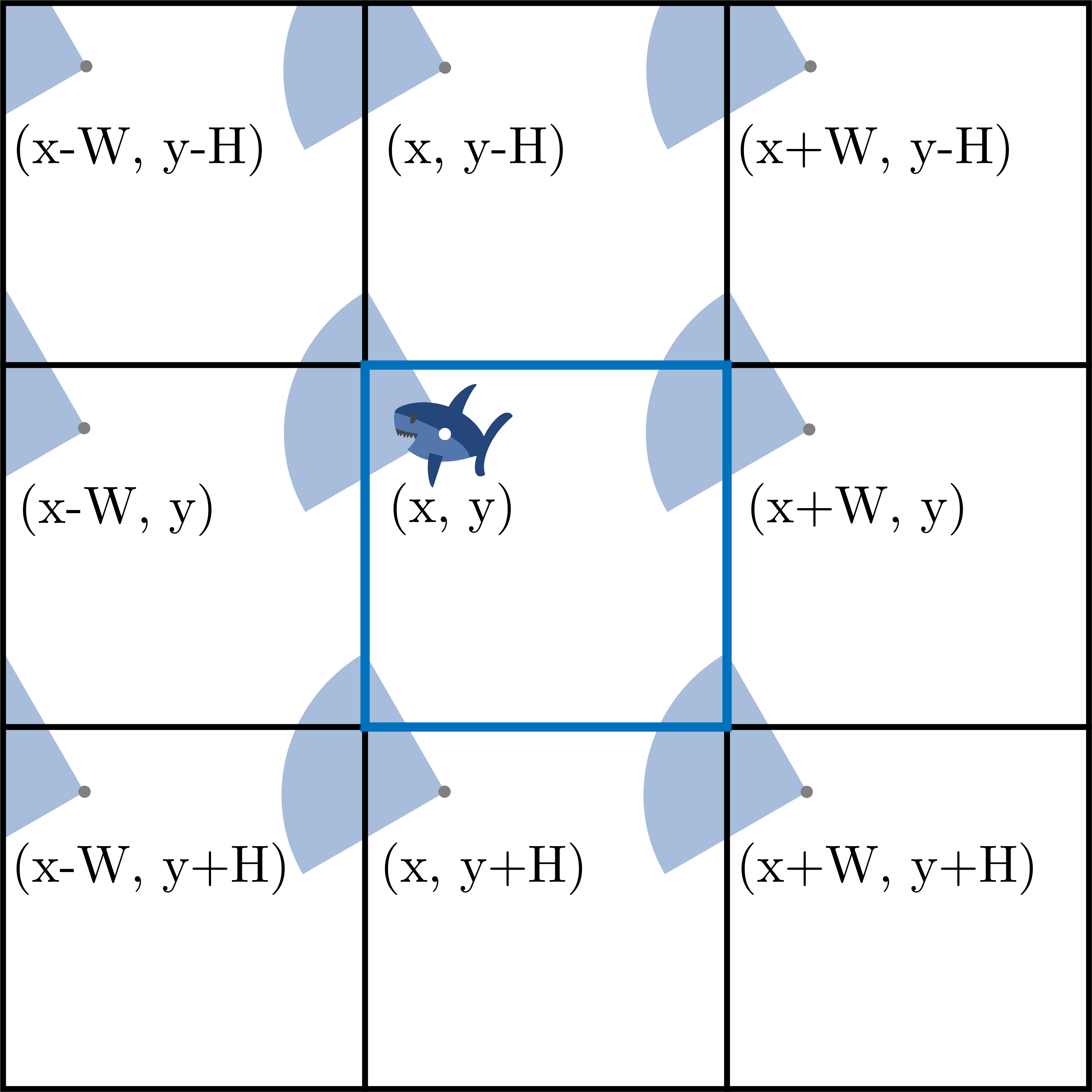}
        \label{fig:cone_torus}
    }
    \hfill
    \subfloat[Distances Between Agents.]{%
        \includegraphics[width=0.47\columnwidth]{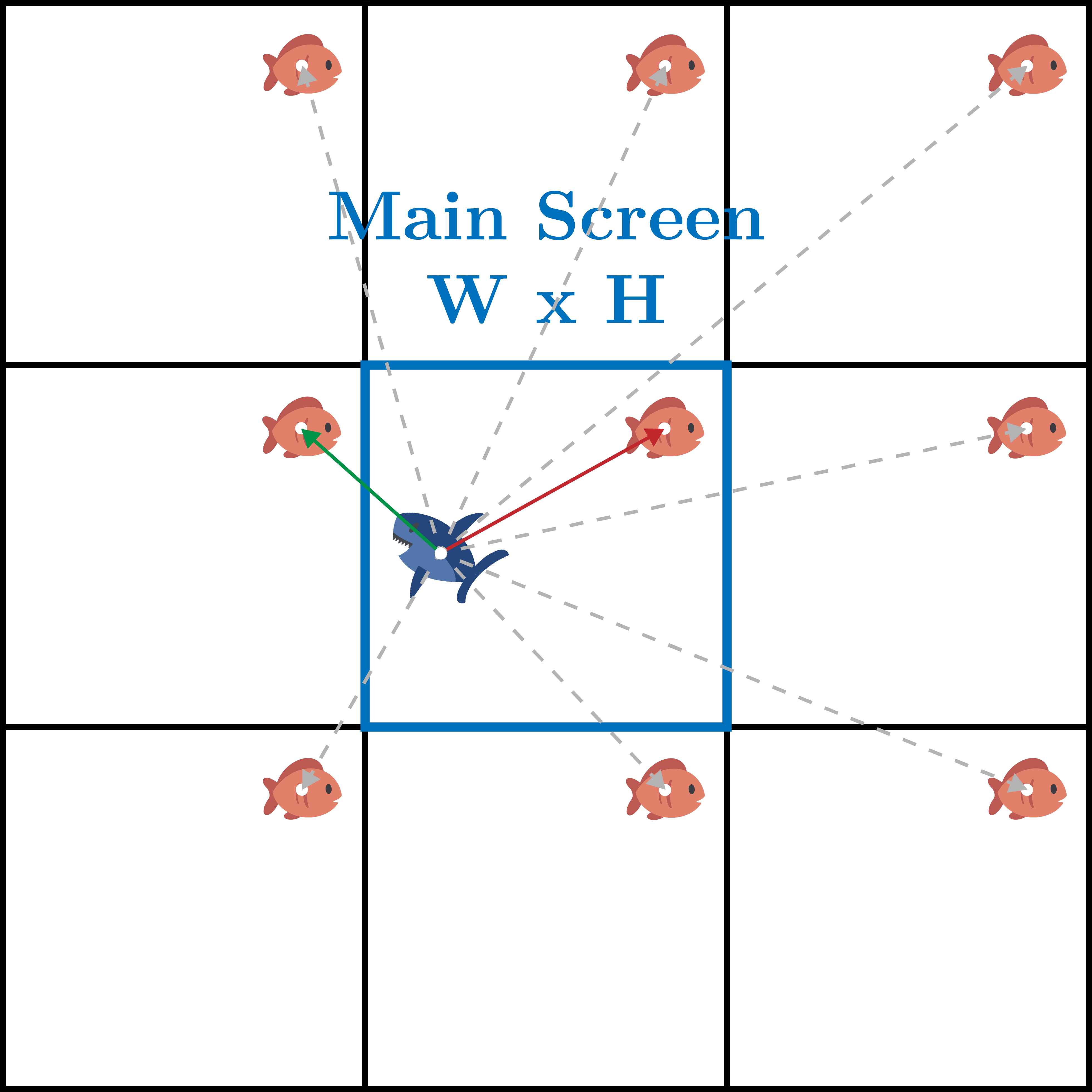}
        \label{fig:intro_torus}
    }
    \caption{Distance and Field of View Calculation in a Torus.}
    \label{fig:torus_figures}
\end{figure}

To further refine the agent's perception, we introduce a restriction, known as the \gls{FOV} constraint, within the agent's frontal direction (\cref{fig:env_example_cone}). This entails that an agent $A$ is only capable of seeing other agents positioned within a specific angular range $\alpha$ in front of it, at a defined distance. As illustrated in \cref{fig:cone_detection}, the determination of an agent's inclusion within its FOV involves fulfilling two specific conditions (\cref{eq:vision}).
\begin{equation}\label{eq:vision}
    \beta \leq \frac{\alpha}{2} \land d \leq v
\end{equation}
Firstly, the angle $\beta$ formed between the agent's current velocity vector $V$ and the desired velocity $D$ must be smaller than $\alpha$. Secondly, the distance $d$ separating the two agents must be shorter than the agent's designated viewing distance $v$. To account for agents wrapping around the boundaries, the FOV is duplicated eight times, each instance situated at distinct locations, as depicted in \cref{fig:cone_torus}. Note that this approach may not be the most efficient approach and there remains the potential for future enhancements. The values for the agent's designated viewing distance $v$ and FOV angle $\alpha$ can be independently adjusted for each agent type in this environment. 

\subsubsection{Distance} \label{sec:distance}
In a toroidal environment, determining the shortest distance between two positions becomes intricate due to the multiple boundary-crossing paths. Unlike a bounded two-dimensional plane, which has a singular distance between points, a torus offers numerous direct paths, some of which may be shorter than conventional distances, as depicted in \cref{fig:intro_torus}. Our primary objective is to find the shortest of these paths. Instead of computing all potential distances and choosing the smallest, we initially assess the standard Euclidean distance within screen space, followed by conditional adjustments based on the environment's width ($W \in \mathbb{N}$) and height ($H \in \mathbb{N}$). Thus, the distance $d$ between two positions $A$ and $B$ can be calculated as:

\begin{equation}\label{eq:distance_in_torus}
d(A,B) = \sqrt{
\begin{split}
&\min(|x_{A}-x_{B}|, W - |x_{A}-x_{B}|)^{2} + \\
&\min(|y_{A}-y_{B}|, H - |y_{A}-y_{B}|)^{2}
\end{split}
}
\end{equation}

\subsubsection{Direction Vector} \label{sec:dir-vec}
To obtain the direction from one point to another in the form of a vector, we introduce two points, $A$ and $B$, representing the initial position and destination. Additionally, let $W \in \mathbb{N}$ and $H \in \mathbb{N}$ be the width and height of the environment. The directional vector is required for the heuristics described in \cref{sec:exp_setup_baselines}. We define the directional vector $D$ with $D(A,B) = (x_D, y_D)$, where

\begin{equation}\label{eq:direction_vector}
\begin{aligned}
x_D =
\begin{cases}
    x_a - x_b, & \text{if } |x_b - x_a| > \frac{W}{2} \\
    x_b - x_a, & \text{otherwise}
\end{cases} \\
\text{where} \\
y_D =
\begin{cases}
    y_a - y_b, & \text{if } |y_b - y_a| > \frac{H}{2} \\
    y_b - y_a, & \text{otherwise}
\end{cases}.
\end{aligned}
\end{equation}

\subsection{Agent-Environment Interaction}\label{sec::ppa_agent_env}

In our modeled ecosystem, the relationship between agents and their environment is crucial for comprehending and influencing predator-prey interactions. Utilizing the Markov Decision Processes framework, we divide the agent behaviors into three key components: observations, actions, and rewards, each of which will be elaborated in the following subsections. It is imperative to note that since our environment operates in a deterministic manner, all transition probabilities are equal to one.

\subsubsection{Observations}

%Observations are the information that each individual learning agent receives from the environment. 
%Due to the environment's classification as partially observed, observations constitute a subset of the overall state of the environment. 
The environment is partially observable and therefore observations constitute only a subset of the overall state of the environment. Both types of agents can possess different observation spaces that can be manipulated through varying configurations. The environment offers the flexibility to use an FOV mechanism characterized by predefined parameters encompassing view distance and angle, as described in \cref{sec:Vision}. Furthermore, a restriction can be imposed on the number of neighboring agents that an agent can perceive. For instance, predators can be restricted to perceive a maximum of three prey agents. Through this modular approach to managing agent observations, the environment provides maximal customizability and reduces coupling between components, enabling the simulations to be more readily adapted to novel research needs.

\begin{figure}[htbp]
    \centering
    \includegraphics[width=\columnwidth]{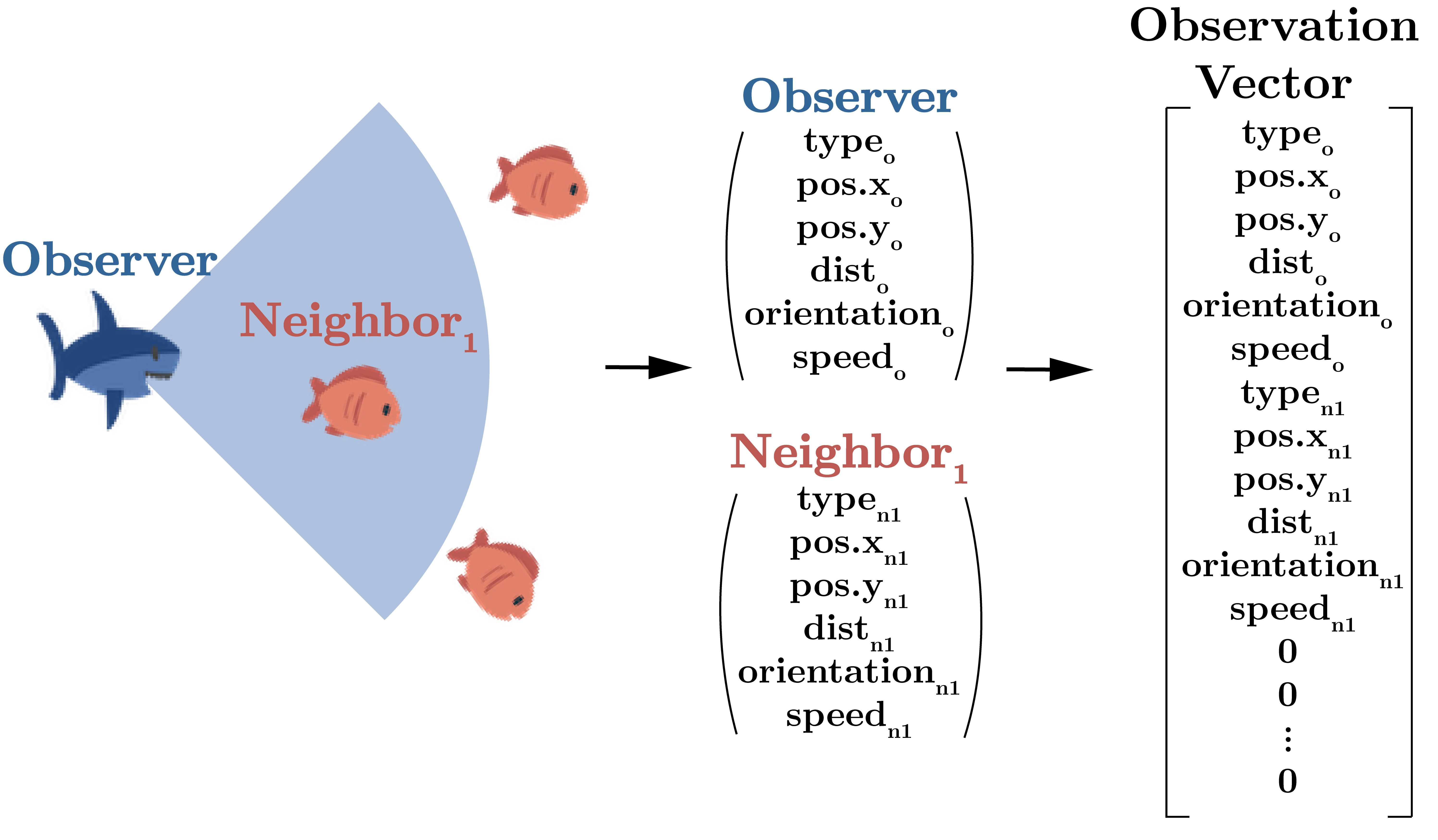}
    \caption{Construction of the Observation Vector. The observation vector is an ordered vector containing 6-tuples for the observer and the $b$ nearest neighboring agents, with $b$ designating the upper limit of observable neighboring agents.}
    \label{fig:observation}
\end{figure}

Under the standard observation configuration, an observing agent $o$ receives a 6-tuple of each neighboring agent $e$, encompassing the neighboring agent's type, position, distance, orientation, and speed (\cref{fig:observation}). The position is represented in terms of polar coordinates, whereas distance, orientation, and speed are characterized by continuous numerical values. Given the environment's toroidal structure, the shortest euclidean distance to the neighboring agent is taken, as described in \cref{sec:distance}. An agent's orientation is quantified in degrees within the range of $[0^\circ, 360^\circ)$, while the speed is constrained by the maximum speed an agent can have. 

Consequently, each observer receives this observation tuple for itself and the $b$ nearest neighboring agents with $b$ representing the upper limit of perceivable neighboring agents. All information is encapsulated within an ordered vector of constant length, restricted by the value of $b$, and within which the $b$ neighbors are ordered based on their respective distance. In scenarios, where an agent's perceptual field yields fewer neighboring agents than the defined maximum $b$, the vector length is preserved by zero padding. Subsequently, this vector undergoes a rescaling procedure to fit within a range spanning from 0 to 1 or alternatively, -1 to 1, depending on the context. This transformation into a vectorized format empowers the interpretation of observations by the \gls{RL} algorithm, enhancing its ability to process the information effectively.

\subsubsection{Actions}

The environment uses distinct modular functions to execute agent actions. This facilitates the customization of independent action spaces for both predators and prey, aligning with the specific requirements of research objectives. In pursuit of broad applicability, the action space within the environment is designed to be compatible with a variety of agent models, including machine learning algorithms, random actions, and static algorithms. Regardless of the specific control logic, all agents share the same defined action space, delineating the permissible movement options within the environment. 

\begin{figure}[htbp]
    \centering
    \includegraphics[width=0.45\columnwidth]{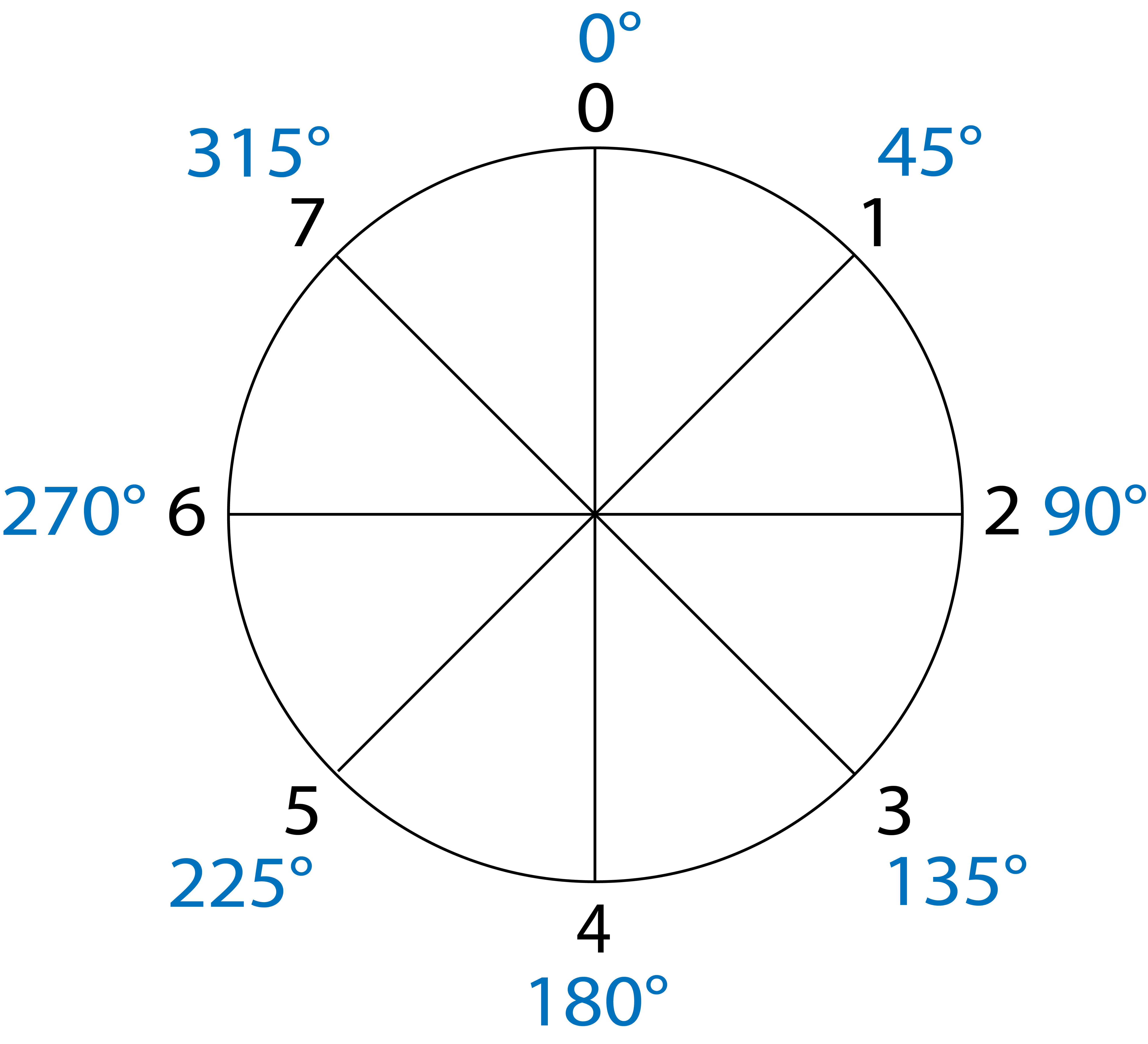}
    \caption{Illustration of the Action Space. Agents select actions from a discrete set of integers, each representing a specific direction. For example, in this 8-option action space, choosing 2 indicates an intent to move east.}
    \label{fig:action_space}
\end{figure}

Within this environment, a discrete action space is utilized, wherein action choices are encoded as integers that correspond to particular directional movements that agents can solicit. For instance, if the action space encompasses eight possible choices, as depicted in \cref{fig:action_space}, an action $a$ of value 0 symbolizes an intention to advance northward, while a value of 1 could denote a northeastern movement, and so forth. These designated directions are subsequently translated into degrees to calculate the desired velocity that dictates the intended agent movement. To turn an action $a$ from an agent's model into an angle of degrees $\gamma$, \cref{eq:action_to_velocity} is applied, with $n$ being the number of actions and desired velocity vector $D$ is then calculated accordingly. 

\begin{equation}\label{eq:action_to_velocity}
    D = \begin{bmatrix} 
    \sin(\gamma \cdot \frac{\pi}{180}) \\ 
    -\cos(\gamma \cdot \frac{\pi}{180}) 
    \end{bmatrix} 
    \quad \text{where} \quad
    \gamma = \frac{360 \cdot a}{n}
\end{equation}

The method by which the desired velocity influences the agent's movement is detailed in \cref{sec:Agent_Movement}.

\subsubsection{Rewards}\label{sec:rewards}

Rewards serve as the primary feedback signal enabling RL algorithms to improve agents' policies. In our configuration, the predator agent receives solely a predefined reward for catching a prey agent, modeling the goal of hunting success. Every other state is classified as neutral, providing neither reward nor punishment, which ensures that no behavioral bias is introduced. In environments featuring multiple predators, the predefined reward $r$ received upon catching a prey can be divided. When there are $n$ predators located within a shared catch zone of a predetermined radius from the prey location, each predator receives a reward of $\frac{r}{n}$.

The goal of the prey is to not collide with a predator and to survive as long as possible. For this, they receive a positive predetermined reward for each time step survived and another negative specified reward for the collision with a predator which ends their life. 

The reward structure for the predator agents was adapted from~\cite{ritz_sustainable_2021}, while the reward structure for the prey agents was inspired by the concepts introduced by~\cite{hahn_emergent_2019}. While the existing setup of the environment relies on these predetermined reward frameworks, it's worth noting that the environment incorporates modular functionalities that enable easy modifications to the established reward system.

\subsection{Limitations}\label{sec:ppa_limitations}

%This environment serves as a foundational platform for \gls*{MARL} predator-prey research, offering essential components such as configurable agent movement, observations, actions, and rewards, functioning as a preliminary research tool.
Despite the environment's adaptability to a variety of scenarios, we sacrifice generality in favor of simplicity by using a two-dimensional plane.
This limits the scope of applications: in nature, predator-prey interactions generally occur in a three-dimensional context.
Also, most UAV scenarios require a third degree of freedom (altitude).
During the time of writing, the environment lacks metrics for cohesion, flocking, and other collective phenomena inherent to predator-prey settings, which we plan to add in future releases.
Also, the environment size is bound by the simulation efficiency, which we plan to improve by optimizing vector operations and the \gls*{FOV} mechanic.
%Ultimately, using Python sacrifices performance in favor of accessibility, but we doubt switching to C++ would increase the environment's adoption given how widely Python is used in academic research.

% This environment enables foundational investigations for \gls*{MARL} predator-prey research by providing key building blocks like realistic movement of configurable agents, observations, actions, and rewards, as an early research tool. However, it sacrifices generality for an adaptable set of core predator-prey problems. Given that real predator-prey scenarios occur in three dimensions, the exclusive use of a two-dimensional environment constrains the variety of studies that can be undertaken. Furthermore, the environment currently has no metric to check for cohesion, flocking behavior, population size, or other natural phenomenon in the predator-prey setting. Moreover, the extent of the environment's size is constrained by the efficiency of the algorithms employed, which have not yet achieved full optimization. Examples of unoptimized algorithms are vector operations, the \gls*{FOV} mechanic, and the fact that the environment is built in Python.

%% file: content/5_experimental_setup.tex
\section{\uppercase{Experimental Setup}} \label{sec:experimental-setup}

We demonstrate the potential of the \textit{Aquarium} environment as a suitable predator-prey scenario for \gls*{MARL} by resembling typical research setups found in related work.
We follow the experimental setup of~\cite{hahn_emergent_2019}, where multiple prey are trained with RL to escape a predator that is guided by a heuristic called \textit{NaivChase}.
\textit{Aquarium} used the default values for all parameters which can be found in the provided repository.
In the following, we describe how we train the prey, which baselines we use and which metrics we choose to evaluate the performance.

\subsection{RL Algorithm}\label{sec:exp_setup_PPO}

We train our prey agents with the RL algorithm \gls*{PPO}~\cite{schulman_PPO_2017} using the \gls*{GAE}.
The hyper-parameters are outlined in \cref{tab:PPO_Hyperparameter}.
We conducted two experiments using an identical parameterization for environment and RL algorithm.
In the first experiment, we use the MARL paradigm \textit{Individual Learning} (IL) to train the prey agents~\cite{schroeder_2020_independent}.
Here, each agent trains an individual policy with its own experiences.
In the second experiment, we use \textit{Parameter Sharing} (PS).
Here, all agents share the parameters of one policy.
Hence, the prey agents learn from experience collected collectively.
For this experiment, the batch size was proportionally reduced by dividing it by the count of agents. For IL and PS, training was performed for $4000$ episodes, each lasting $3000$ time steps with five different seeds.

\begin{table}[!h]
\small
\centering
\begin{tabularx}{\linewidth}{Xcc}
\toprule
Parameter & Data Type & Value \\
\midrule
Discount Factor & Float & 0.99 \\
Batch Size & Integer & 2048 \\
Clipping Range & Float & 0.1\\
GAE Lambda & Float & 0.95\\
Entropy Weight & Float & 0.001\\
Actor Alpha & Float & 0.001\\
Critic Alpha & Float & 0.003\\
\bottomrule
\end{tabularx}
\caption[PPO Hyper-Parameters]{PPO Hyper-Parameters. These hyper-parameters were used for training the prey agents in an environment including a single predator agent.}
\label{tab:PPO_Hyperparameter}
\end{table}

\subsection{Baselines}\label{sec:exp_setup_baselines}

To asses the trained RL policies, we implemented several heuristic baselines.

The \textbf{Random} heuristic operates by selecting actions in an arbitrary manner, devoid of any strategic consideration or learning process.
This represents an untrained agent and any training should result in a significant improvement.

The \textbf{Static} heuristic is a different set of rules for predator and prey agents.
In case of prey agents, the heuristic applies the rules of the \textbf{TurnAway} algorithm reported by~\cite{hahn_emergent_2019}, where agents turn $180\degree$ away from the predator.
This involves computing the directional vector from the prey to the predator (see \cref{sec:dir-vec}), inverting it and converting it into an angle.
Then, this angle is mapped to an action within the action space (\cref{fig:action_space}).
In case of predator agents, the heuristic applies the rules of the \textbf{NaivChase} algorithm reported by~\cite{hahn_emergent_2019}.
It determines the direction vector from the predator to the closest prey and converts it into an action.
If a predator has multiple prey within its FOV, it arbitrarily selects one prey to pursue at each time step, modelling confusion.

\subsection{Metrics}\label{sec:metrics}

To measure the learning success and analyze the population dynamics, our framework provides two pivotal metrics.

\textbf{Rewards} is the undiscounted sum of all rewards collected per episode.
This metric measures how well an agent is doing on the long run.
It does not have to be normalized since the episode length is constant.
Predator agents are rewarded for capturing prey and prey agents are rewarded for surviving,
Higher values indicate better performance.

\textbf{Captures} is the sum of captured prey per episode.
It measures the success of predators agents catching prey (from this point of view, higher values are better) and the success of prey agents evading the predators (from this point of view, lower values are better).
Prey agents reappear after being caught to keep the episode length constant.

%% file: content/6_results.tex
\section{RESULTS} \label{sec:results}

The following section encompasses the outcomes of the two experiments elucidated in \cref{sec:exp_setup_PPO}. Initially, the efficacy of the training approach wherein individual agents possess distinct policies is juxtaposed with the baseline models described in \cref{sec:idv_training}. Subsequently, a comparison between the two experiments employing distinct training strategies is presented in \cref{sec:idv_training_vs_param}. The trained policies were executed across 200 episodes, each consisting of 3000 time steps, using five distinct seeds in an environment featuring a single predator governed by the \textit{NaivChase} heuristic. The same protocol was followed for the baseline models. During these runs, both undiscounted rewards and captures per episode were collected for each prey agent to compare the different models on their evading performance. The results of the different seeds were summarized by averaging the respective metrics.

\subsection{Individual Learning}\label{sec:idv_training}

To recapitulate, each prey agent developed its own distinct policy through dedicated training, ensuring that their policy's learning process exclusively derived from their unique set of experiences. Our aim is to replicate the findings observed in the study conducted by~\cite{hahn_emergent_2019}. While we did not integrate a specific metric to assess the extent of prey cooperation, such as swarming, we anticipate discovering that the fish learn to enhance their survival by maintaining a distance from the predator. Nevertheless, we do not anticipate the prey agent to embrace the action of executing a $180^\circ$ turn away from the predator, as performed by the \textit{TurnAway} heuristic.

\begin{figure}[htbp]
    \centering
    \includegraphics[width=\columnwidth]{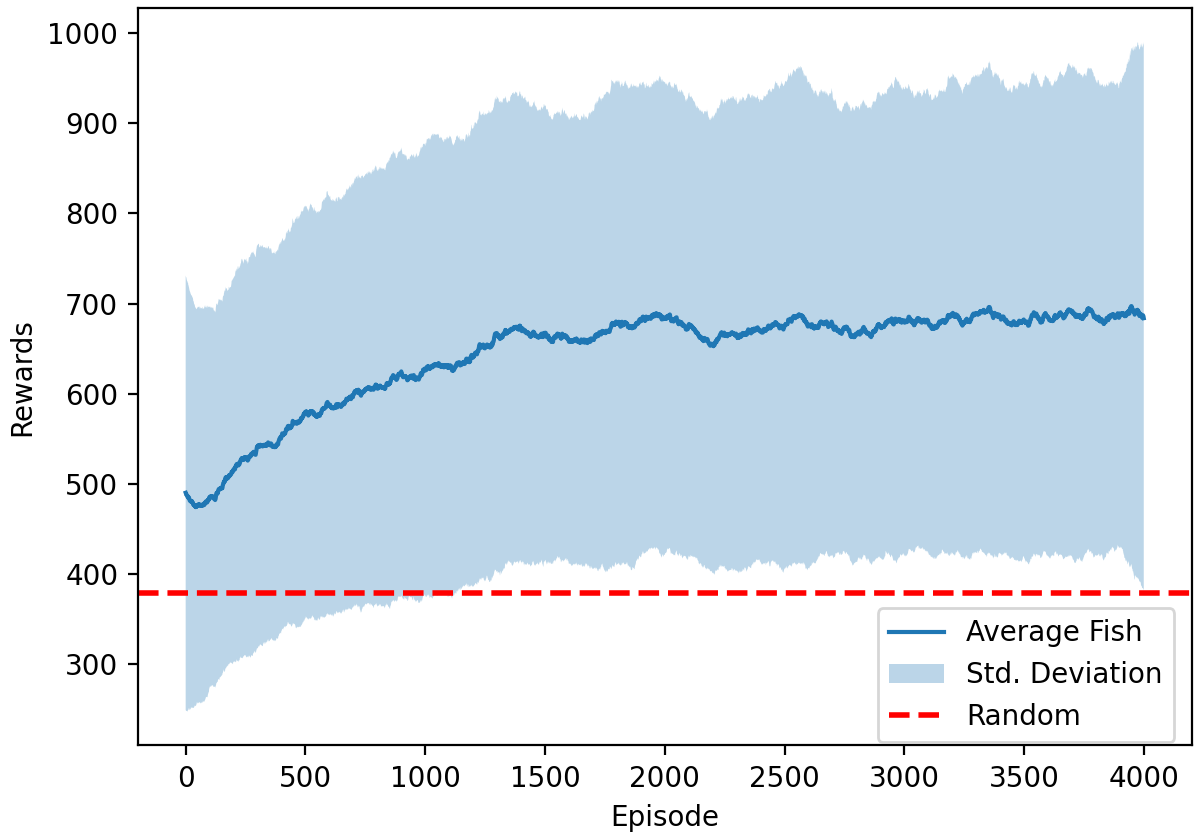}
    \caption[Average Reward per Prey Agent Using the Individual Training Strategy.]{Average Reward per Prey Agent Using the Individual Training Strategy. Training was performed for $4000$ episodes, each lasting $3000$ time steps. The prey agents were individually trained, such that each agent is equipped with its unique policy and exclusively learns from its own experiences. The individual rewards over the six prey agents were averaged for each episode and the standard deviation was calculated. The red dotted line represents the average reward achieved overall episodes with random behavior.}
    \label{fig:indi_rewards}
\end{figure}

First, we examined the rewards and capture per episode during the training process. Initially, the policy’s average rewards, calculated over all prey agents, closely align with that of the random policy, as illustrated in \cref{fig:indi_rewards}. Subsequently, there is a rapid growth in reward, and after approximately 2000 episodes the increase becomes gradual. The standard deviation reveals a pronounced dispersion of rewards across the entire training.

Upon examining the captures per episode, a similar trend becomes apparent. Initially, the policy's capture outcomes closely resemble those of the random policy. Nonetheless, these captures undergo a substantial decline until around episode $2000$. After this point, the capture rate demonstrates only a slight additional reduction. Once more, the standard deviation exhibits notable elevation, suggesting considerable variability across episodes regarding the frequency of prey capture events by the predator.

\begin{figure}[htbp]
    \centering
    \includegraphics[width=\columnwidth]{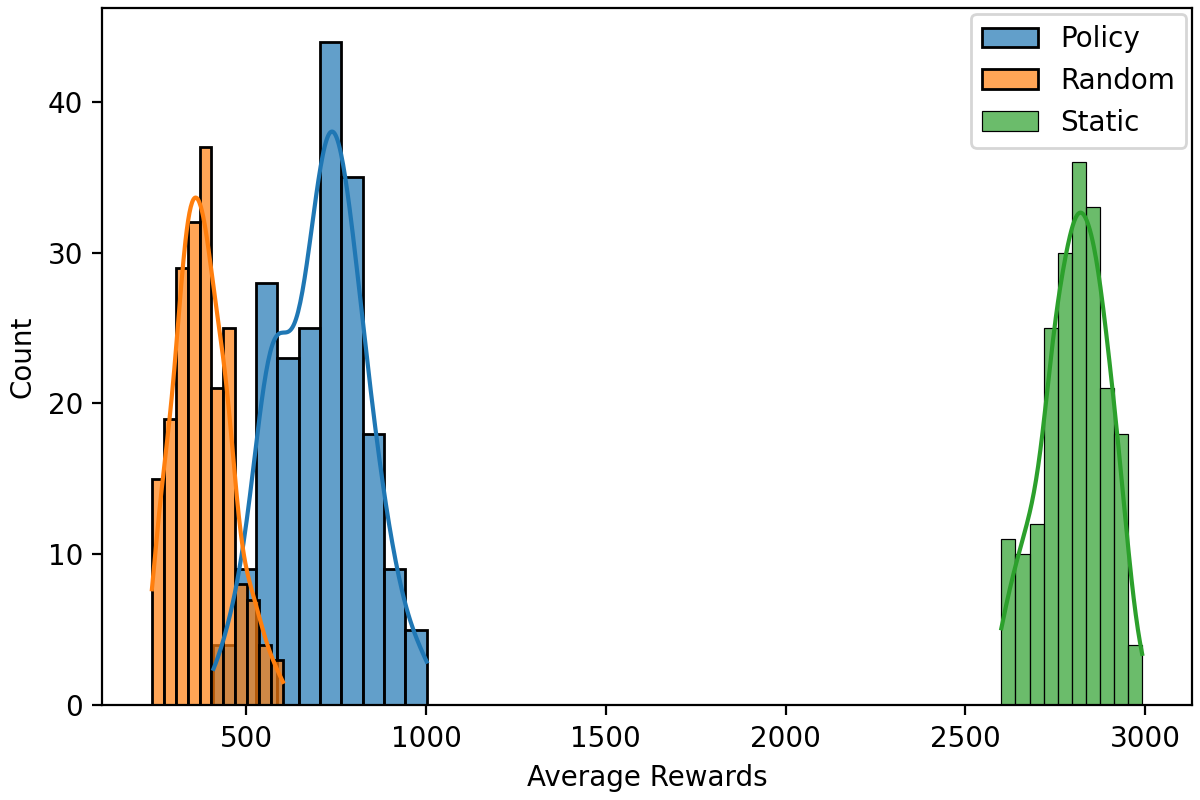}
    \caption[Distributions of the Average Rewards of Three Different Models.]{Distribution of the preys' average rewards in a scenario with one predator. We compared the trained RL prey agent (blue) against the \textit{Random} heuristic (orange) and the \textit{TurnAway} heuristic (green) across five differnent seeds.}
    \label{fig:indi_rewards_combined}
\end{figure}

When comparing the behavior of the prey agents using the learned policies to the behavior of those being based on two baselines model, observed across a span of 200 episodes, it becomes evident that the \textit{TurnAway} heuristic has pronounced effectiveness, yielding notably higher average rewards in comparison to both the random and trained prey models. Nonetheless, it is imperative to acknowledge that the trained policy consistently maintains superior performance over the random policy in relation to the specified reward metric. Upon evaluating the captures per episode across the three models, the prey subjected to the trained policy exhibited an average capture frequency of approximately five times per episode (\cref{fig:indi_rewards_combined}). In contrast, the prey, under the influence of the static algorithm, experienced a notably lower capture rate, with fewer than one capture per episode on average. This underscores the effectiveness of the static algorithm in countering the predator with similar control.

%In addition to the two computed evaluation metrics, a video file was recorded and is available under \href{https://drive.google.com/file/d/1X8uJdkac5hGkOfg3WhYg8xhJahyaRPas/view?usp=sharing}{\nolinkurl{https://drive.google.com/file/d/1X8uJdkac5hGkOfg3WhYg8xhJahyaRPas/view?usp=sharing}}. 
The trained prey agents exhibited a slight tendency to move in a direction opposite to that of the predator. However, their effectiveness in executing this evasion behavior remained limited, as they persisted in directly engaging in actions aimed at the predator. This choice of action resulted in a substantial number of captures for the prey, indicating that the evasion strategy is not highly successful in preventing capture incidents.

Overall, the prey agents individually trained using the PPO technique exhibit a marginal improvement over random movement patterns. However, their performance remained notably inferior when contrasted with the efficacy demonstrated by the \textit{TurnAway} heuristic. This observation aligns closely with the findings of~\cite{hahn_emergent_2019}, albeit with the distinction that a more comprehensive set of metrics is required to comprehensively assess complex behaviors such as swarming.

\subsection{Parameter Sharing}\label{sec:idv_training_vs_param}

\begin{figure}[htbp]
    \centering
     \subfloat[Average Captures Over Prey Agents.]{%
        \includegraphics[width=0.47\columnwidth]{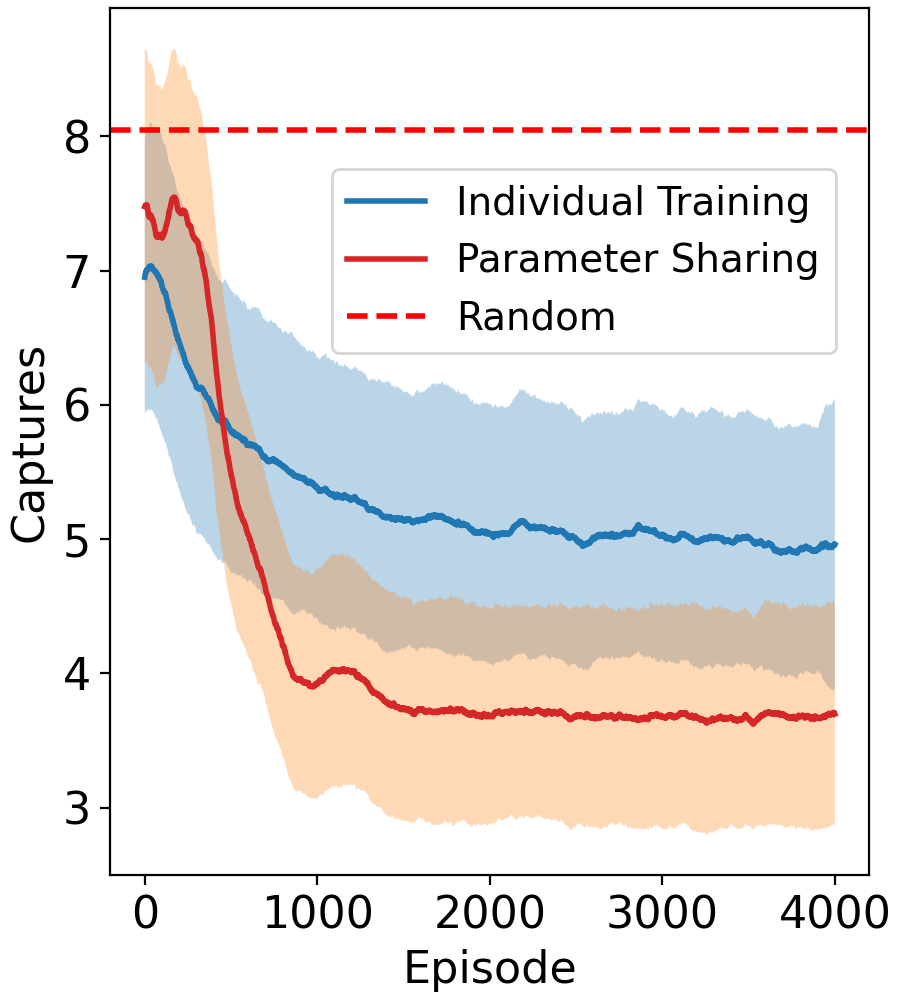}
        \label{fig:comparison_line_capture}
    }
    \hfill
    \subfloat[Average Rewards Over Prey Agents.]{%
        \includegraphics[width=0.47\columnwidth]{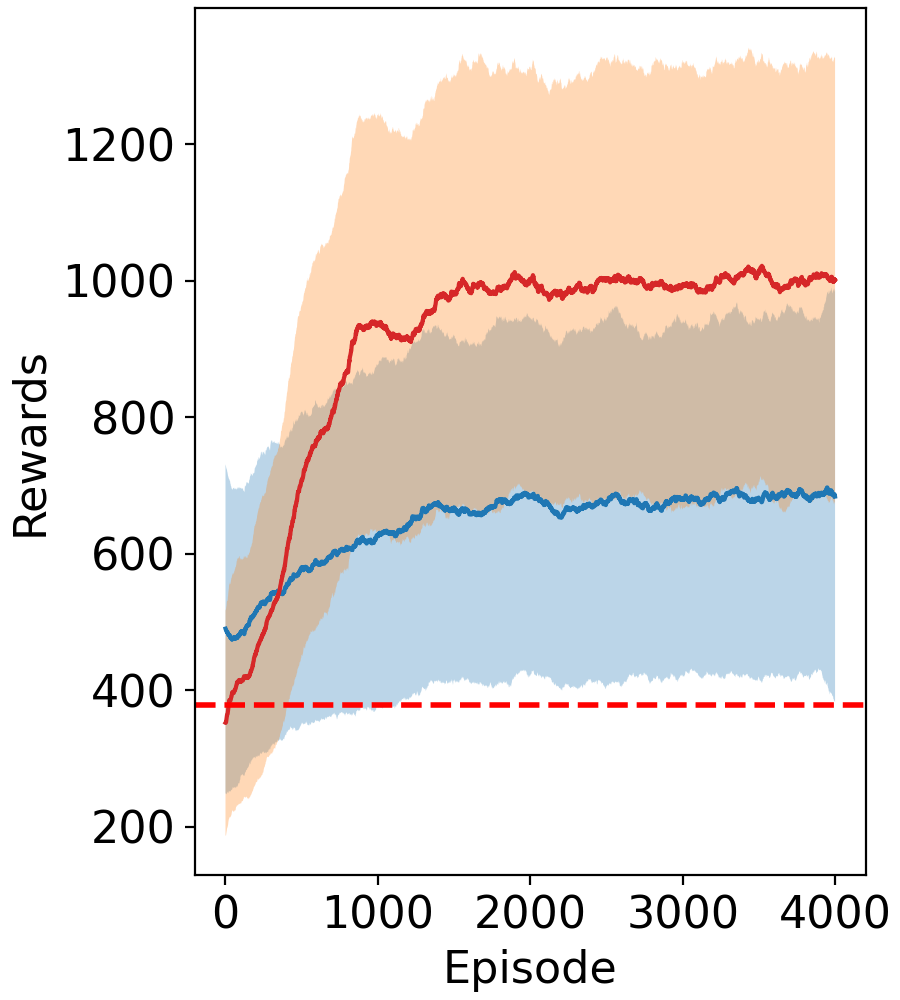}
        \label{fig:comparison_line_reward}
    }
    \hfill
    \vspace{5pt}
    \caption[Average Captures and Rewards per Prey Agent Comparing Both Training Strategies.]{Average captures and rewards per prey agent. We compare individual learning (blue) against parameter sharing (red). Training was performed for $4000$ episodes, each lasting $3000$ time steps. The metrics are averaged over six prey agents and five distinct seeds per episode. The shaded areas represent the respective standard deviation.}
    \label{fig:comparison_line}
\end{figure}

Utilizing parameter sharing among prey agents, as discussed in \cref{sec:exp_setup_PPO}, means that each agent benefits from the collective experiences of the group, learning from a single policy source. This approach is expected to accelerate the learning rate, potentially leading to cooperative behaviors or swarm formations, confusing the predator (\cite{hahn_emergent_2019}). Our preliminary results suggest enhanced survival rates for prey, as they strategically maintain distance from predators, indicated by the rapid decline in captures per episode (\cref{fig:comparison_line_capture}) and the parallel increase in rewards (\cref{fig:comparison_line_reward}).

\begin{figure}[htbp]
    \centering
     \subfloat[Average Captures Over Prey Agents.]{%
        \includegraphics[width=0.47\columnwidth]{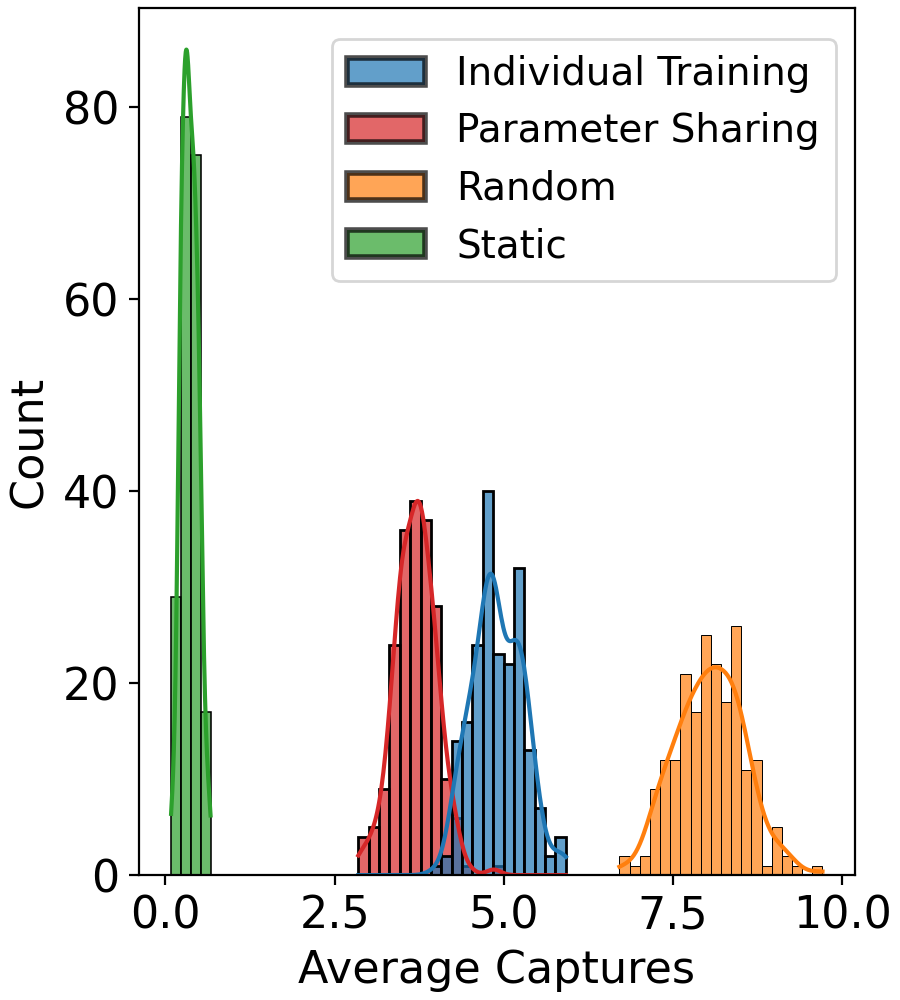}
        \label{fig:comparison_hist_capture}
    }
    \hfill
    \subfloat[Average Rewards Over Prey Agents.]{%
        \includegraphics[width=0.47\columnwidth]{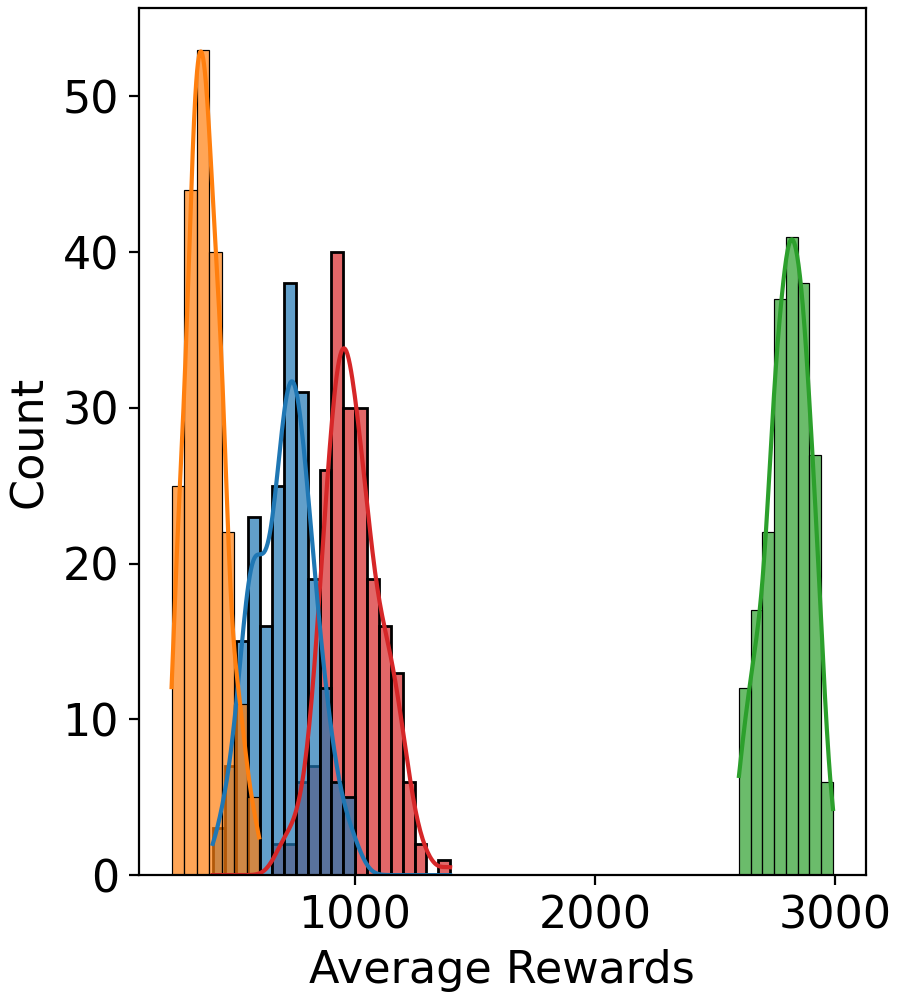}
        \label{fig:comparison_hist_reward}
    }
    \hfill
    \vspace{5pt}
    \caption[Distributions of the Average Captures and Rewards Comparing Both Training Strategies.]{Distributions of the Average Captures and Rewards Comparing Both Training Strategies. The trained policies, individual training (blue) and training with parameter sharing, the random heuristic (orange), and the \textit{TurnAway} heuristic (green) were executed across 200 episodes, each consisting of 3000 time steps, using five distinct seeds in an environment featuring a single predator governed by the \textit{NaivChase} heuristic.}
    \label{fig:comparison_hist}
\end{figure}

Additionally, the prey employing the parameter sharing approach demonstrate markedly improved post-training performance, aligning with our initial anticipation of accelerated learning. The prey subjected to parameter sharing training outperform the random agents, exhibiting reduced capture rates and superior rewards compared to the individually trained prey (\cref{fig:comparison_hist}). Nevertheless, the performance of prey governed by the \textit{TurnAway} heuristic still surpasses that of the other models.

\begin{figure}[!h]
    \centering
    \includegraphics[width=\columnwidth]{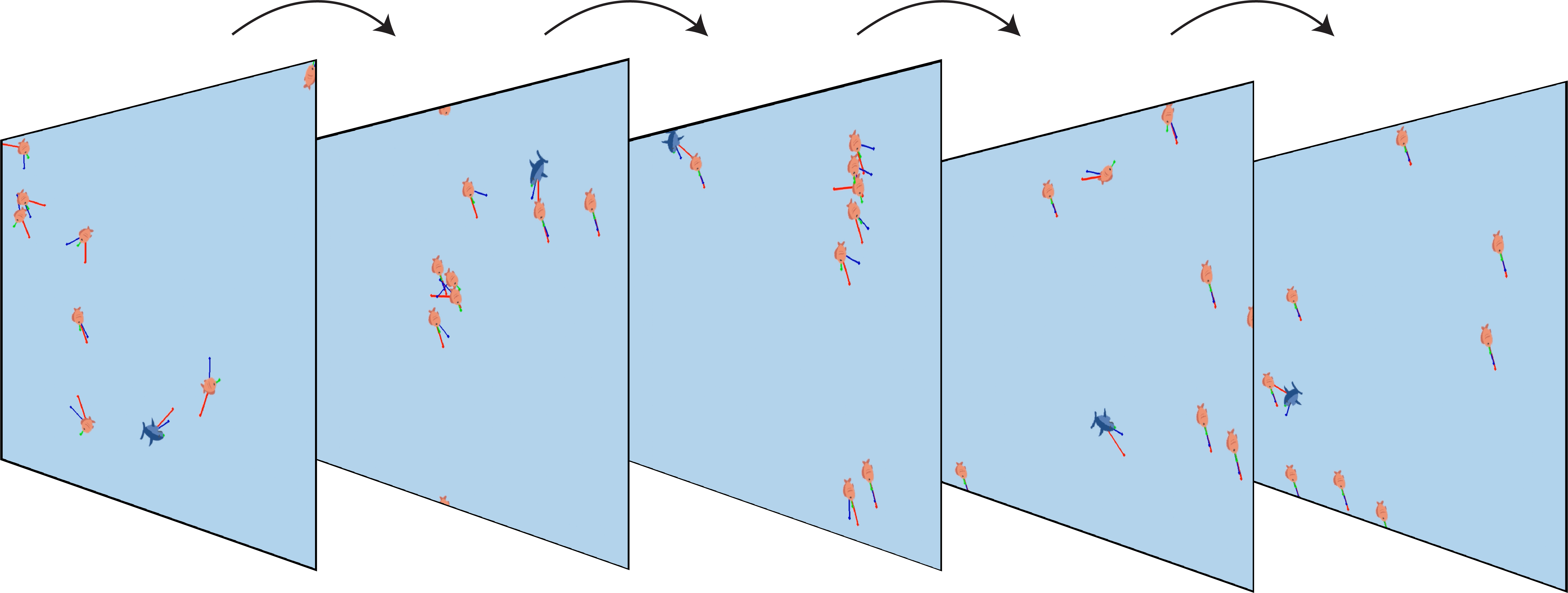}
    \caption[Illustration of Directional Movement Evolution of Prey Agents Using Training With Parameter Sharing.]{Illustration of Directional Movement Evolution of Prey Agents Using Training With Parameter Sharing. Training was performed for 600 episodes, each lasting $3000$ time steps. A single policy was trained collectively based on the observations of all prey agents. To illustrate the behavior of the prey agents, five screenshots at an interval of 100 episodes were recorded.}
    \label{fig:evolution}
\end{figure}

Upon examining the recorded video of one episode of the environment consisting of prey agents trained using parameter sharing, 
%accessible at \href{https://drive.google.com/file/d/15qYcIy5KdXhjZMV7TYrUZ-q2cjW8AJEn/view?usp=sharing
%}{\nolinkurl{https://drive.google.com/file/d/15qYcIy5KdXhjZMV7TYrUZ-q2cjW8AJEn/view?usp=sharing
%}}, 
we noticed that the policy guides all agents to consistently swim in the same direction. The progress of this phenomenon becomes evident during training. \cref{fig:evolution} showcases five distinct images captured at various time steps throughout the training process. This observation was unexpected, as this straightforward behavior of maintaining uniform direction appears primitive in attempting to evade the predator. However, it can be deduced that the \gls*{PPO} algorithm converged to a local optimum. Remarkably, this aligns with the adoption of the alignment rule of swarm behavior, as elucidated in~\cite{reynolds_flocks_1987}, which ultimately granted them an advantage over the individually trained, self-centered prey agents.

%% file: content/7_conclusion.tex
\section{\uppercase{Conclusion}} \label{sec:conclusion}
% Intro
In this work, we introduced \textit{Aquarium}, a comprehensive and flexible MARL environment that models predator-prey interaction.
By providing an overview of existing predator-prey environments, we identified key aspects required by the (MA)RL community.
Based on that, we provide a customizable implementation that covers all identified aspects and is compatible to the proven MARL algorithm implementations of the \textit{PettingZoo} framework~\cite{terry_pettingzoo_2021}.
In preliminary experiments, we reproduced emergent behaviour of learning agents and demonstrated the scalability of modern MARL paradigms in our environment.

Future prospects can be divided into three categories: improving the environment implementation, adding further features and conducting comprehensive experiments.
Regarding the environment implementation, we hope to reduce the computational footprint in various aspects, such as optimizing vector operations and streamlining computations required for the utilization of the FOV mechanism.
In particular, the integration of ray tracing techniques~\cite{raytracing} has the potential to substantially improve performance.
We believe that the pivotal expansion of the environment to accommodate a larger number of agents is imperative for the in-depth analysis of swarm behavior.
Regarding additional features, we plan to add vector flow fields~\cite{reynolds1999steering} that simulate water flow or wind.
This would allow to investigate how agents behave in presence of external forces. 
Regarding experiments, we plan to replicate group hunting as reported by~\cite{ritz_sustainable_2021} and explore the (optional) FOV mechanism, e.g. to test the \textit{Many Eye Hypothesis}~\cite{olson_2015_vigilance} which has not received much attention yet.
Ultimately, we hope for the community to adopt our environment and provide feedback on deficiencies we may have overlooked.

% End
%In conclusion, the Aquarium stands as a flexible research platform, designed for the investigation of MARL in predator-prey scenarios.
%Through the implementation of a framework that adjusts to a diverse range of scenarios using numerous adaptable parameters, it not only establishes a favorable backdrop for in-depth studies but also optimizes the efficiency of executing and contrasting various experiments, and readily facilitates the replication of previous experiments.

%% file: content/8_acknowledgements.tex
\section*{ACKNOWLEDGEMENTS}
This work is part of the Munich Quantum Valley, which is supported by the Bavarian state government with funds from the Hightech Agenda Bayern Plus.